\documentclass[sigconf]{acmart}

\usepackage{bbm}
\usepackage{algorithm}
\usepackage{algorithmic}
\usepackage{bbding}
\usepackage{fancyhdr}
\AtBeginDocument{%
  \providecommand\BibTeX{{%
    \normalfont B\kern-0.5em{\scshape i\kern-0.25em b}\kern-0.8em\TeX}}}

\setcopyright{acmcopyright}
\copyrightyear{2021}
\acmYear{2021}
\setcopyright{acmcopyright}\acmConference[MM '21]{Proceedings of the 29th ACM International Conference on Multimedia}{October 20--24, 2021}{Virtual Event, China}
\acmBooktitle{Proceedings of the 29th ACM International Conference on Multimedia (MM '21), October 20--24, 2021, Virtual Event, China}
\acmPrice{15.00}
\acmDOI{10.1145/3474085.3475622}
\acmISBN{978-1-4503-8651-7/21/10}
\acmSubmissionID{mfp2486}

\begin{document}
\fancyhead{}
\title[Co-learning]{Co-learning: Learning from Noisy Labels with Self-supervision}

%  EMPTY

%  EMPTY

\author{
    Cheng Tan$^{1,2}$, Jun Xia$^{1,2}$, Lirong Wu$^{1,2}$, Stan Z. Li$^{1,2}$
}\authornote{Corresponding author.}
\affiliation{
    $^{1}$\institution{AI Lab, School of Engineering, Westlake University.}
    $^{2}$\institution{Institute of Advanced Technology, Westlake Institute for
Advanced Study}
\institution{18 Shilongshan Road, Hangzhou 310024, Zhejiang Province, China}
    \country{}
}
\email{{tancheng,xiajun,wulirong,stan.zq.li}@westlake.edu.cn}

% \renewcommand{\shortauthors}{Anonymous Author, et al.}

\begin{abstract}
Noisy labels, resulting from mistakes in manual labeling or webly data collecting for supervised learning, can cause neural networks to overfit the misleading information and degrade the generalization performance. Self-supervised learning works in the absence of labels and thus eliminates the negative impact of noisy labels. Motivated by co-training with both supervised learning view and self-supervised learning view, we propose a simple yet effective method called \emph{Co-learning} for learning with noisy labels. Co-learning performs supervised learning and self-supervised learning in a cooperative way. The constraints of intrinsic similarity with the self-supervised module and the structural similarity with the noisily-supervised module are imposed on a shared common feature encoder to regularize the network to maximize the agreement between the two constraints. Co-learning is compared with peer methods on corrupted data from benchmark datasets fairly, and extensive results are provided which demonstrate that Co-learning is superior to many state-of-the-art approaches. 
The code is available at \href{https://github.com/chengtan9907/Co-training-based\_noisy-label-learning}{GitHub}.

\end{abstract}

\begin{CCSXML}
<ccs2012>
   <concept>
       <concept_id>10010147.10010178.10010224</concept_id>
       <concept_desc>Computing methodologies~Computer vision</concept_desc>
       <concept_significance>500</concept_significance>
       </concept>
 </ccs2012>
\end{CCSXML}
\ccsdesc[500]{Computing methodologies~Computer vision}

\keywords{label noise, self-supervised learning, computer vision}

\maketitle

\section{Introduction}

The success of deep learning relies on carefully labeled data in large-scale. However, precise annotations are extremely expensive and time-consuming. To alleviate the problem, inexpensive alternatives are often used, such as web crawling \cite{thomee2015new} or completing annotations with crowd-sourcing and online queries. Unfortunately, these alternative methods can often leads to noisy labels. However, deep neural networks can easily overfit to noisy labels, as shown in recent research \cite{zhang2017understanding,mahajan2018exploring,li2019learning,tanno2019learning}, and this can dramatically degrade the generalization performance.

\begin{figure}[htbp]
\centering
\includegraphics[width=0.46\textwidth]{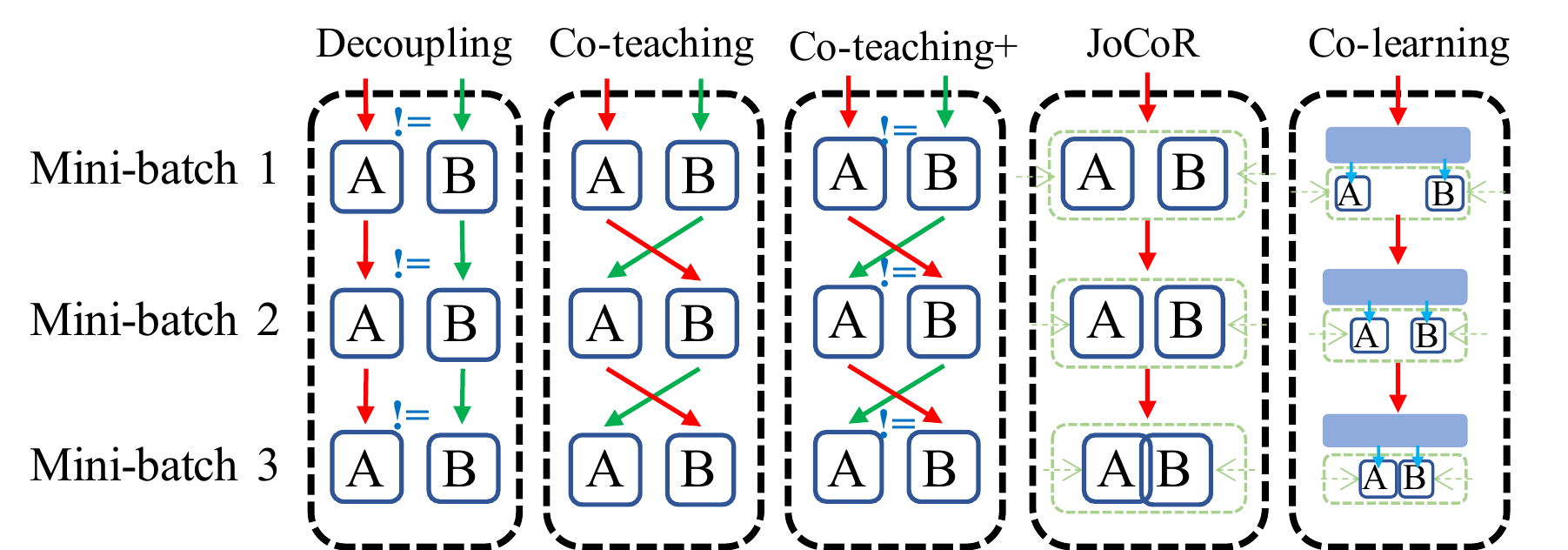} 
\caption{Comparison of leading methods in dealing with noisy labels. All the methods use two cooperative modules (A \& B). As the training goes, the two become more and more in agreement with each other, but in different ways. 
 \textit{Decoupling} updates the parameters of the two networks with prediction-disagreed (!=) samples from a mini-batch.
 \textit{Co-teaching} \cite{han2018co} uses small-loss samples of one network to teach its peer network for further training. \textit{Co-teaching+} \cite{yu2019does} first predicts each mini-batch with the two networks but uses disagreed (!=) samples only to compute the training loss. 
 \textit{JoCoR} \cite{wei2020combating} trains the two networks as a whole with a joint loss of weighted goals: making the two predictions agree with each other, and making the predictions stick to ground-true labels as far as possible. 
 \textit{Co-learning} trains a single shared encoder network with two heads (the self-supervised and the noisily-supervised) that constrain each other and maximizes the agreement between them in latent space.}
\label{fig:comparison} 
\end{figure}

Several studies have been conducted to investigate learning with noisy labels, where semi-supervised learning frameworks play crucial roles \cite{yan2016robust, ding2018semi, nguyen2019self, arazo2019unsupervised, li2019dividemix}. Most of these deploy unsupervised learning to obtain information about label-independent features and then do further training using noisy labels in a supervised learning manner. These methods commonly leverage labels through two views: (1) supervised learning taking advantage of direct supervision with labels and (2) unsupervised learning exploiting intrinsic features from data to combat label noise. The core challenge here is how to combine the two views effectively. Arazo et al. \cite{arazo2019unsupervised} design a beta mixture model as an unsupervised generative model of sample loss values and then adopt MixUp \cite{zhang2017mixup} augmentation to assure reliable convergence under extreme noise levels. Li et al. \cite{li2019dividemix} models sample loss with Gaussian mixtures to divide training data into unlabeled data and labeled data, and then apply the state-of-the-art semi-supervised learning method MixMatch \cite{berthelot2019mixmatch} to deal with the divided sets. However, the above methods are not end-to-end and the step-by-step learning manner increases complexity in both time and space.

Co-training and model ensembling have been shown to be beneficial in learning with noisy labels \cite{malach2017decoupling, han2018co, yu2019does, wei2020combating}. Decoupling \cite{malach2017decoupling} trains two networks simultaneously and updates them using instances with different predictions. Co-teaching \cite{han2018co} also trains two networks simultaneously and selects small-loss data to teach the peer network during the training process. Co-teaching+ \cite{yu2019does} follows a similar scheme as Co-teaching but with a scheme which selects small-loss data from disagreement data. JoCoR \cite{wei2020combating} maintains two networks but trains them as a whole with a joint loss to make predictions of them get closer. These methods are based on the assumption that two networks can provide two different views of the data, but the extra information gain they bring is very limited since the differences between two networks of the same architecture mainly come from random initialization. A comparison between Co-learning and other co-training-like methods is illustrated in Figure~\ref{fig:comparison}.

Motivated by semi-supervised learning and co-training, we propose a simple yet effective learning paradigm called "\emph{Co-learning}" to combat the problem with noisy labels, in which self-supervised learning is introduced as a feature-dependent view to assist supervised learning. Different from those co-training-like methods, Co-learning has a shared feature encoder with two exclusive heads. While the projection head performs a self-supervised learning and exploits feature-dependent information through the intrinsic similarity, the classifier head performs a vanilla supervised learning and learns from label-dependent information. In addition, a structural similarity loss is utilized between the two exclusive heads to regularize the classifier and avoid bias caused by noisy labels. This is based on the constraint that  output from the projection head and that from the classifier head should share similar structure pairwise. Unlike the other methods, Co-learning can be implemented conveniently without the need for such prior knowledge as noise rates, data distributions, and additional clean samples, thus avoiding the needed hyperparameters thereof. 

We conducted experiments on simulated and real-world noisy datasets, including CIFAR-10, CIFAR-100, Animal-10N \cite{song2019selfie}, and Food-101N \cite{lee2018cleannet} datasets. Among these datasets, Animal-10N is a noisy dataset that contains confusing images for manual annotations, and Food-101N is a webly dataset that directly collects data from the Web. Comparative results demonstrate that Co-learning is superior to state-of-the-art methods and robust to high-level noise in labels.
\section{Related work}

\subsection{Learning with noisy labels}

Various methods have been studied towards robust learning with noisy labels, and we briefly review existing works on learning with noisy labels in this section.
\paragraph{Noise transition matrix} Some of the existing methods estimate the label transition matrix. Forward-correction \cite{patrini2017making} performs forward correction by multiplying the estimated transition matrix with the softmax outputs during the forward propagation step. Gold loss correction \cite{hendrycks2018using} leverages available trusted labels for loss correction and utilizes trusted labels' confusion matrices to obtain an accurate transition matrix. However, estimating the transition matrix is challenging in real-world settings and maybe infeasible for finding closed-form solutions. 
\paragraph{Sample selection} Another direction of handling noisy labels without noise transition matrix is training on selected samples. These approaches try to select "clean" instances according to certain rules. In particular, MentorNet \cite{jiang2018mentornet} pretrains a teacher network for selecting clean instances to guide the training of a student network. Co-teaching \cite{han2018co} chooses clean instances from a network to feed its peer network for updating parameters. Co-teaching+ \cite{yu2019does} and JoCoR \cite{wei2020combating} also highlight the small-loss sample selection to explicitly assist the co-training procedures of two networks. INCV \cite{chen2019understanding} divides noisy training data randomly to apply cross-validation and removes large-loss samples. This family of methods avoids the risk of error correction by simply excluding unreliable samples. Although these methods have achieved significant improvements, they neglect the potential supervisory signal of the large-loss training samples. Besides, either the noise rate or a clean validation set is required as prior knowledge. 
\paragraph{Semi-supervised learning} Several recent studies transform the problem of learning with noisy labels into a semi-supervised learning task. There are mainly two directions in these methods: (1) divide noisy training data into two sets where reliable samples are treated as labeled data, and unreliable samples are treated as unlabeled data, then perform certain existing semi-supervised learning methods \cite{yan2016robust, ding2018semi, arazo2019unsupervised, li2019dividemix, nguyen2019self}, (2) perform unsupervised learning on noisy training data to assist the supervised learning \cite{han2019deep, mandal2020novel}. Our method follows the second direction that utilizes unsupervised learning to implicitly help to learn from noisy labels.

\subsection{Self-supervised learning}

\begin{figure}[htbp]
\centering
\includegraphics[width=0.46\textwidth]{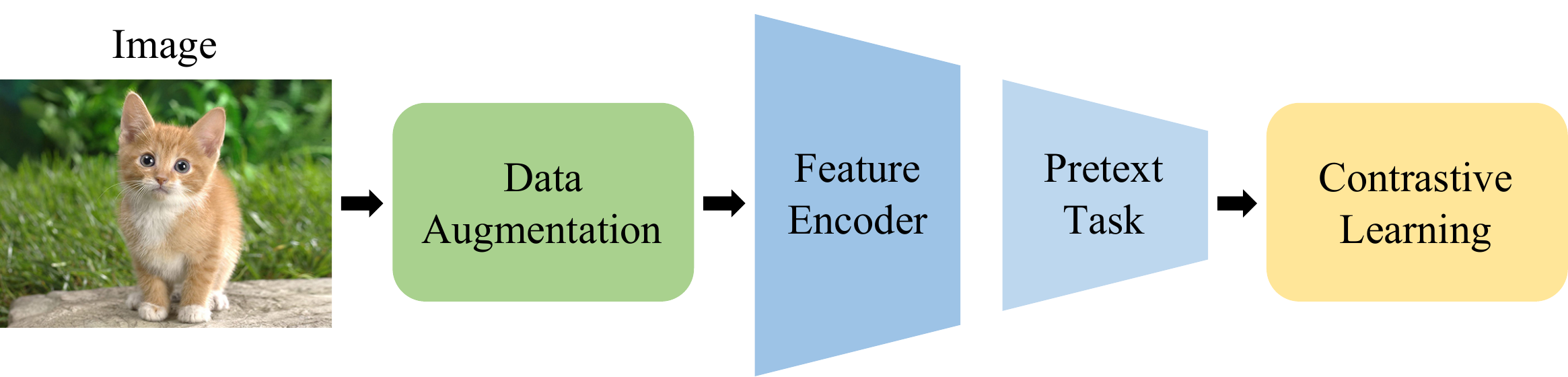} 
\caption{Typical contrastive learning pipeline.}
\label{fig:contrastive_learning} 
\end{figure}

Self-supervised learning is a subset of unsupervised learning that designs pretext tasks to produce labels derived from the data itself. While solving pretext tasks, the model learns valuable representations; that is, the feature encoder of the model is able to extract useful features. There are various pretext tasks that can help the model learn representations, such as colorization \cite{zhang2016colorful}, inpainting \cite{pathak2016context}, rotation \cite{gidaris2018unsupervised}, jigsaw \cite{noroozi2016unsupervised} and contrastive learning \cite{he2020momentum, chen2020simple, grill2020bootstrap, chen2020exploring, caron2020unsupervised}. Among them, recent studies on contrastive learning have become a dominant component in self-supervised learning methods. Contrastive learning aims at grouping similar samples closer and diverse samples away from each other. A typical contrastive learning pipeline is shown in Figure \ref{fig:contrastive_learning}. To measure the similarity between the feature representations, contrastive learning utilizes a metric that we call \emph{intrinsic similarity} in this paper.

Current state-of-the-art contrastive learning methods mainly rely on a combination of intrinsic similarity and a set of image transformations. MoCo~\cite{he2020momentum} builds a dynamic dictionary with a queue and a momentum encoder to obtain large and consistent dictionaries of visual representations. SimCLR~\cite{chen2020simple} shows that the composition of data augmentations is crucial and a nonlinear projection head will further improve the quality of the learned representations. While contrastive learning may fail into collapsed solutions like outputting the same representations for all samples, BYOL~\cite{grill2020bootstrap} takes advantage of both the target network and the slow-moving average online network to avoid collapsing. To simplify the framework in BYOL, SimSiam~\cite{chen2020exploring} proposes Siamese networks with a stop-gradient operation. SwAV~\cite{caron2020unsupervised} introduces an online clustering strategy to further improve the performance without either a large memory bank or a momentum encoder.

\subsection{Relations to other approaches}

We compare Co-learning with related methods and illustrate the main differences in Table \ref{tab:comparison}. Although Co-learning is motivated by Co-training and other Co-training-like methods \cite{malach2017decoupling, han2018co, yu2019does, wei2020combating}, the only similarity is that Co-learning tries to maximize agreements from different views.

\begin{table}[ht]\footnotesize
\centering
\begin{tabular}{lcccc}
\toprule
                   & Co-teaching & Co-teaching+ & JoCoR & Co-learning \\
\midrule
Agreement          & \Checkmark & \Checkmark & \Checkmark & \Checkmark \\
Small-loss         & \Checkmark & \Checkmark & \Checkmark & \XSolidBrush \\
Double classifiers & \Checkmark & \Checkmark & \Checkmark & \XSolidBrush \\
Cross update       & \Checkmark & \Checkmark & \XSolidBrush & \XSolidBrush \\
\bottomrule
\end{tabular}
\caption{Comparison with state-of-the-art methods in techniques used -- “Agreement": maximizing the agreement from different views by regularization; "Small-loss": regarding small-loss samples as “Clean” samples; “Double classifiers”: training two classifiers simultaneously; “Cross update”: updating parameters in a cross manner instead of a parallel manner.}
\label{tab:comparison}
\end{table}

These methods treat the samples with small losses as reliable data and abandon the remaining samples. They cannot take full advantage of all the given data, hindering the potential of the model. This small-loss trick deliberately avoided the problem of difficult samples and noisy samples. Furthermore, their small-loss trick needs noisy rates, and other carefully chosen hyperparameters \cite{han2018co, yu2019does, wei2020combating} lead to difficulties in real-world applications. 

Our approach approves the assumption that different views can give more reliable information on noisy training data. However, the above methods simply apply two same classifiers that can provide little variety on views. The different views from networks with the same architecture rely on random initialization. Instead of maximizing agreements on models' predictions, Co-learning tries to maximize agreements on the feature encode $f$ so that both feature-dependent and label-dependent information can play roles during the training procedure. A shared feature encoder in Co-learning not only ensures the soft agreements from different views but also reduces computation cost to a great extent.

\section{Methods}
\subsection{Preliminaries}
Considering a classification problem with $c$ classes, suppose $\mathcal{X} \subset \mathbb{R}^d$ is the input space and $\mathcal{Y}=\{0,1\}^c$ is the ground-truth label space in an one-hot manner. The training set $\mathcal{D}=\{(x^{(i)},y^{(i)})\}_{i=1}^{N}$ is obtained from a joint distribution over $\mathcal{X}\times\mathcal{Y}$. For the general supervised learning, the model learns a mapping function $\mathcal{F}:\mathcal{X} \to \mathcal{Y}$ to minimize the empirical risk $\mathcal{R}_{\mathcal{L}}(\mathcal{F})$ under a loss function $\mathcal{L}$,
\begin{equation}
\mathcal{R}_{\mathcal{L}}(\mathcal{F}) = \mathbb{E}_{\mathcal{D}}\left [\mathcal{L}(\mathcal{F}(x),y)\right] = \frac{1}{\lvert\mathcal{D}\rvert}\sum_{(x,y)\in\mathcal{D}}\mathcal{L}(\mathcal{F}(x),y).
\end{equation}

In our case, the mapping function $\mathcal{F}$ is a neural network model parameterized by $\Theta$. The optimal parameters $\Theta^*$ minimize the empirical risk subjected to network parameters,

\begin{equation}
    \Theta^* = \arg \min_{\Theta} \mathcal{R}_{\mathcal{L}}(\mathcal{F}(\;\cdot \;;\Theta))
\end{equation}

A noisy training dataset $\widehat{\mathcal{D}} = \{(x^{(i)}, \hat{y}^{(i)})\}_{i=1}^{N}$ is provided from a noisy joint distribution over $\mathcal{X}\times\mathcal{\widehat{Y}}$ where $\hat{y}_i$ is a noisy label. Under the standard training procedure, a mini-batch $\mathcal{B}_t = \{(x^{(i)}, \hat{y}^{(i)})\}_{i=1}^b$ is randomly selected from $\widehat{\mathcal{D}}$ at time $t$ with batch size $b$. In this condition, parameters of the model may be mistakenly updated according to the gradient direction of the mini-batch $\mathcal{B}_t$,
\begin{equation}
\Theta_{t+1} = \Theta_t - \eta \nabla (\frac{1}{|\mathcal{B}_t|} \sum_{(x, \hat{y}) \in \mathcal{B}_t} \mathcal{L}(\mathcal{F}(x; \Theta_t), \hat{y})),
\end{equation}
where $\eta$ is the learning rate. 

The optimization with noisy labels may easily lead to inaccurate direction, thus misguide and degenerate the model generalization performance on unseen data. The goal of learning with noisy labels is to mitigate the mischief of noisy labels.

\subsection{Co-learning}

\begin{figure}[htbp]
\centering
\includegraphics[width=0.44\textwidth]{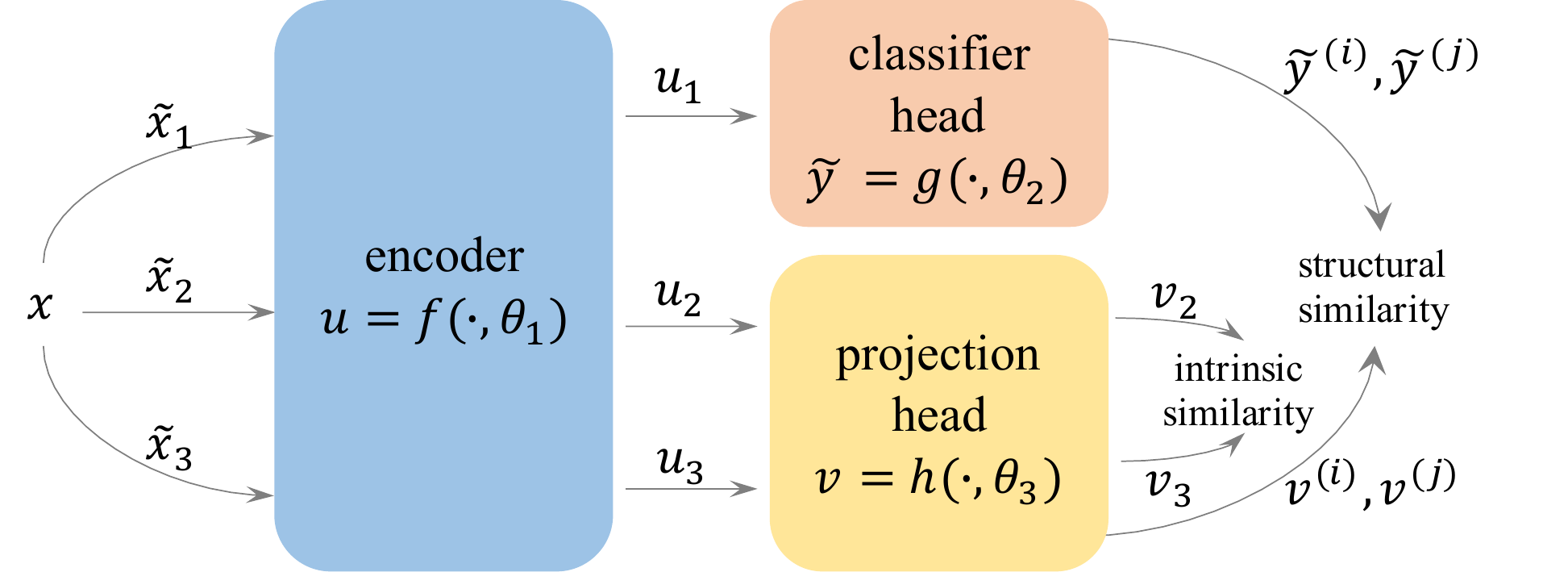} 
\caption{The architecture of Co-learning. The main task of supervised classifier learning and the auxiliary pretext task of self-supervised learning work cooperatively in the process of learning the encoder and the classifier. The subscripts 1, 2 and 3 index the three transformed versions of training example while the superscripts $(i)$ and $(j)$ index pairs of training examples.}
\label{fig:architecture} 
\end{figure}

The Co-learning framework is illustrated in Figure~\ref{fig:architecture}. The main learning task is to learn the encoder and the classifier, which is done with a vanilla cross-entropy loss defined on label-dependent information. A contrastive prediction-based pretext task is used to assist the main task with an {\em intrinsic similarity} loss defined on feature-dependent information, as motivated by the recent success of self-supervised learning \cite{he2020momentum, chen2020simple, grill2020bootstrap, chen2020exploring, caron2020unsupervised}.
A {\em structural similarity} loss is introduced in this work to maximize agreement on the feature encoder shared by the two tasks. We also hope this can help the classifier learn with noisy labels.

Given a mini-batch $\mathcal{B}_t = \{(x^{(i)}, \hat{y}^{(i)})\}_{i=1}^b$, we sample two independent strong transformations from the same distribution $\mathcal{T}$ and one weak transformation from another distribution $\mathcal{T}'$ to obtain the transformed data $\{\Tilde{x}^{(i)}_k\}_{k=1, 2, 3}$, for which the definitions of the strong and weak transformations are illustrated in Appendix. A feature encoder $f$ with parameters $\theta_1$ takes the transformed images as inputs and produces representation vectors $u_k^{(i)}= f(\Tilde{x}^{(i)}_k;\theta_1)$. The first representation $u_1^{(i)}$ is fed into the classifier $g$ with parameters $\theta_2$ and formulated as $\Tilde{y}^{(i)} = g(u_1^{(i)}; \theta_2)$. The projection head $h$ which is a $\theta_3$ parameterized multi-layer perceptron then 
non-linearly transforms $u_2^{(i)}, u_3^{(i)}$ into projections, i.e., $v^{(i)}_2 = h(v_2^{(i)}; \theta_3)$, $v^{(i)}_3 = h(v_3^{(i)}; \theta_3)$.

The intrinsic similarity loss is calculated between $v^{(i)}_2$ and $v^{(i)}_3$ by taking $(v^{(i)}_2, v^{(i)}_3)$ as a positive pair and $\{(v^{(i)}_2, v^{(j)}_2), (v^{(i)}_3, v^{(j)}_3)\}_{j \neq i}$ as the negative pairs. This indicates the same data has intrinsic similarity through different transformations. The two different heads which perform contrastive prediction and supervised classification simultaneously can be regarded as two different pretext tasks. This indicates the same data has structural similarity through different pretext tasks.

Specifically, the complete loss function $\mathcal{L}$ is a linear combination of the supervised loss $\mathcal{L}_{sup}$, intrinsic similarity loss $\mathcal{L}_{int}$, and structural similarity loss $\mathcal{L}_{str}$ as follows:

\begin{equation}
    \mathcal{L} = \mathcal{L}_{sup} + \mathcal{L}_{int} + \mathcal{L}_{str}
\end{equation}

\noindent{\bf Supervised classifier loss.} 
For multi-class classification problems, the cross-entropy loss is a common supervised loss to measure the relative entropy between the corrupted label $\hat{y}$ and its corresponding prediction $\Tilde{y}$. 

\begin{equation}
    \mathcal{L}_{ce}(\hat{y}, \Tilde{y}) = - \sum_{i=1}^N  \hat{y}^{(i)} \log(\Tilde{y}^{(i)})
\end{equation}

Considering contrastive learning requires a long training period comparing to its supervised counterpart, Co-learning deploys MixUp \cite{zhang2017mixup} augmentation to slow down the convergence of the feature encoder $f$ and avoid the supervised learning quickly overfitting to noisy labels. Then we create new training samples by linearly interpolating a sample with another sample (indexed by $m(i)$) randomly chosen from the same mini-batch:

\begin{equation}
    \bar{x}^{(i)} = \lambda \Tilde{x}^{(i)} + (1 - \lambda) \Tilde{x}^{(m(i))}
\end{equation}
\begin{equation}
    \bar{y}^{(i)} = \lambda \hat{y}^{(i)} + (1 - \lambda) \hat{y}^{(m(i))}
\end{equation}
where $\lambda \sim Beta(\alpha, \alpha)$.

Thus, the MixUp version of the supervised loss is formulated as:

\begin{equation}
    \mathcal{L}_{sup} = - \sum_{i=1}^N \left[ \lambda \bar{y}^{(i)} \log(\Tilde{y}^{(i)}) + (1 - \lambda) \bar{y}^{(i)} \log(\Tilde{y}^{(m(i))}) \right]
\end{equation}

\noindent{\bf Intrinsic similarity loss.} 
In the contrastive prediction task, projections of the same image from different views should be similar. The intrinsic similarity between 
$v_2^{(i)}$ and $v_3^{(i)}$ (the projections of the two strongly transformed versions $\Tilde{x}_2^{(i)}$ and $\Tilde{x}_3^{(i)}$ from $x^{(i)}$) is measured by the contrastive loss function as in InfoNCE~\cite{oord2018representation}:

\begin{equation} %\changeto 1&2
    \ell(v_a^{(i)},v_b^{(i)}) = -\log\frac{\exp(D(v_a^{(i)}, v_b^{(i)}) / \tau)}{\sum_{j=1,i \neq j}^{N} \sum_{t_i, t_j \in \{2, 3\}} \exp(D(v_{t_i}^{(i)}, v_{t_j}^{(j)}) / \tau)}
\end{equation}
where $D(v_a, v_b) = \frac{v_a^T v_b}{\|v_a\| \|v_b\|}$ is the pairwise similarity between $v_a$ and $v_b$, and $\tau$ is a temperature parameter. The intrinsic similarity loss is thus defined as:
\begin{equation}
    \mathcal{L}_{int} =  \sum_{i=1}^N \ell(v_2^{(i)}, v_3^{(i)}) + \ell(v_3^{(i)}, v_2^{(i)})
\end{equation}

\begin{figure}[htbp]
\centering
\includegraphics[width=0.48\textwidth]{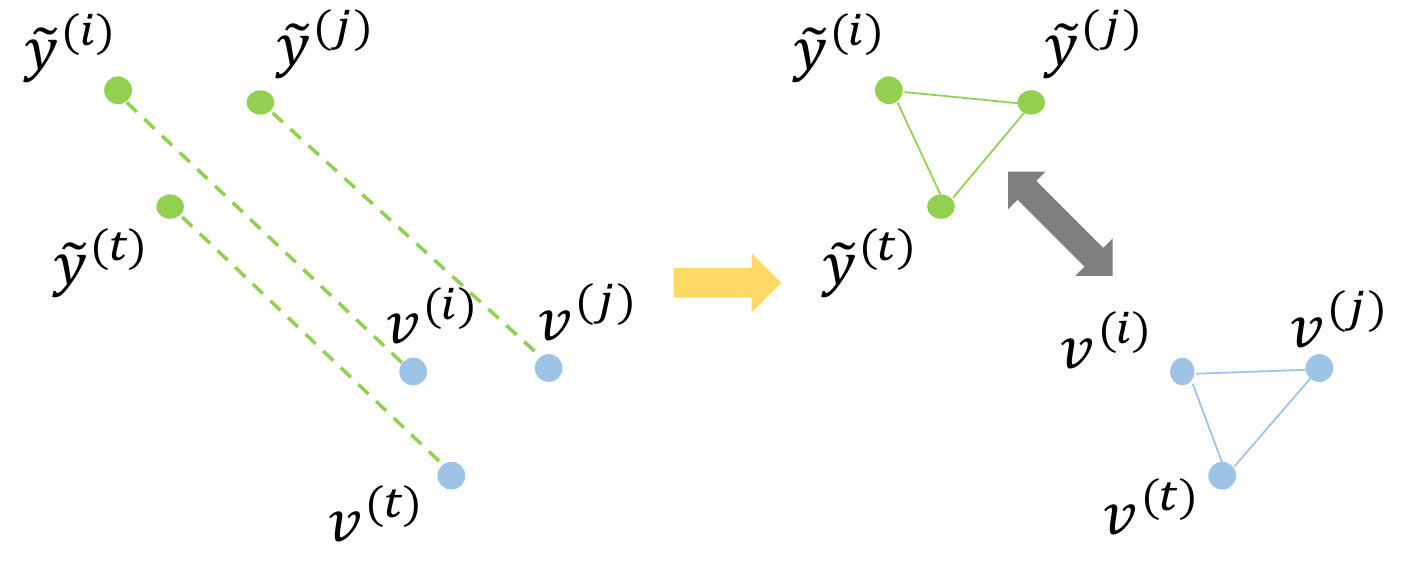} 
\caption{The corresponding outputs from $g$ and $h$ should have similar relational structures between pairs of points. Left: transformation $\tilde{y} \rightarrow v$; Right: The relationship between transformed points is similar to the previous.}
\label{fig:structure} 
\end{figure}

\noindent{\bf Structural similarity loss.}
The classifier $g$ and the projection head $h$ share the same encoder network, so they jointly influence the process of the encoder learning. However, the classifier could be misled by noisy labels. To overcome this problem, we impose a structure-preserving constraint ~\cite{li_MLDL_2020,li_DMT_2020} on pairs of features from $g(\cdot)$ and the corresponding ones from $h(\cdot)$, as illustrated in Figure \ref{fig:structure}. The structure similarity loss is defined as follows. 

First, we convert the Euclidean distances $d(v^{(i)}, v^{(j)})$, $d(\Tilde{y}^{(i)}, \Tilde{y}^{(j)})$ into the similarity metrics $p(d(v^{(i)}, v^{(j)})), p(d(\Tilde{y}^{(i)}, \Tilde{y}^{(j)})) \in [0, 1]$ that satisfy $\lim_{d \rightarrow +\infty} p(d) = 0$ and $\lim_{d \rightarrow 0} p(d) = 1$. The similarity metric is formulated as:

\begin{equation}
    p(d) = C_{\sigma} \frac{1}{\sigma \sqrt{2 \pi}} e^{-\frac{1}{2}(\frac{d - \mu}{\sigma})^2}
\end{equation}
where $C_{\sigma} = \sigma \sqrt{2 \pi}$ is a constant normalizing the similarity metric in $[0, 1]$, and we set $\mu = 0, \sigma = 0.5$ empirically.

Then the structural similarity loss is defined as the KL-divergence between similarity metrics converted from the projection $z$ and logits $\Tilde{y}$:

\begin{equation}
    \mathcal{L}_{str} = \sum_{i \neq j} p(d(v^{(i)}, v^{(j)})) \log \frac{p(d(v^{(i)}, v^{(j)}))}{p(d(\Tilde{y}^{(i)}, \Tilde{y}^{(j)}))}
\end{equation}

The proposed approach is delineated in Algorithm 1.

\begin{algorithm}
    \caption{Co-learning}
    \begin{algorithmic}[1]
        \REQUIRE Encoder $f$ with $\theta_1$, classifier $g$ with $\theta_2$, projection head $h$ with $\theta_3$, learning rate $\eta$, batch size $B$, epoch $T_{\mathrm{max}}$
        \FOR{$t = 1, 2, ..., T_{\mathrm{max}}$}
            \STATE Shuffle training set $\widehat{\mathcal{D}}$
            \FOR{$n = 1, 2, ..., \frac{|\widehat{\mathcal{D}}|}{B}$}
                \STATE Fetch $n$-th mini-batch $\widehat{\mathcal{D}}_n$ from $\widehat{\mathcal{D}}$
                \FOR{$i = 1, 2, ..., |B|$}
                    \STATE Weak augment $\Tilde{x}^{(i)}_1 = t'(x^{(i)})$
                    \STATE Strong augment $\Tilde{x}^{(i)}_2 = t_1(x^{(i)}), \Tilde{x}^{(i)}_3 = t_2(x^{(i)})$
                    \STATE Obtain representations $u_k = f(\Tilde{x}_k^{(i)}; \theta_1), k \in \{1, 2, 3\}$
                    \STATE Obtain prediction $\Tilde{y} = g(u_1; \theta_2)$
                    \STATE Obtain projections $v_k = h(u_k; \theta_3), k \in \{2, 3\}$
                \ENDFOR
                \STATE Calculate the loss $\mathcal{L}$ by Eq. (4)
                \STATE Update $\Theta = \Theta - \eta \nabla \mathcal{L}$
            \ENDFOR
        \ENDFOR
        \RETURN $\theta_1$, $\theta_2$, $\theta_3$
    \end{algorithmic}
\end{algorithm}

\begin{figure*}[t]
\centering
\includegraphics[width=\textwidth]{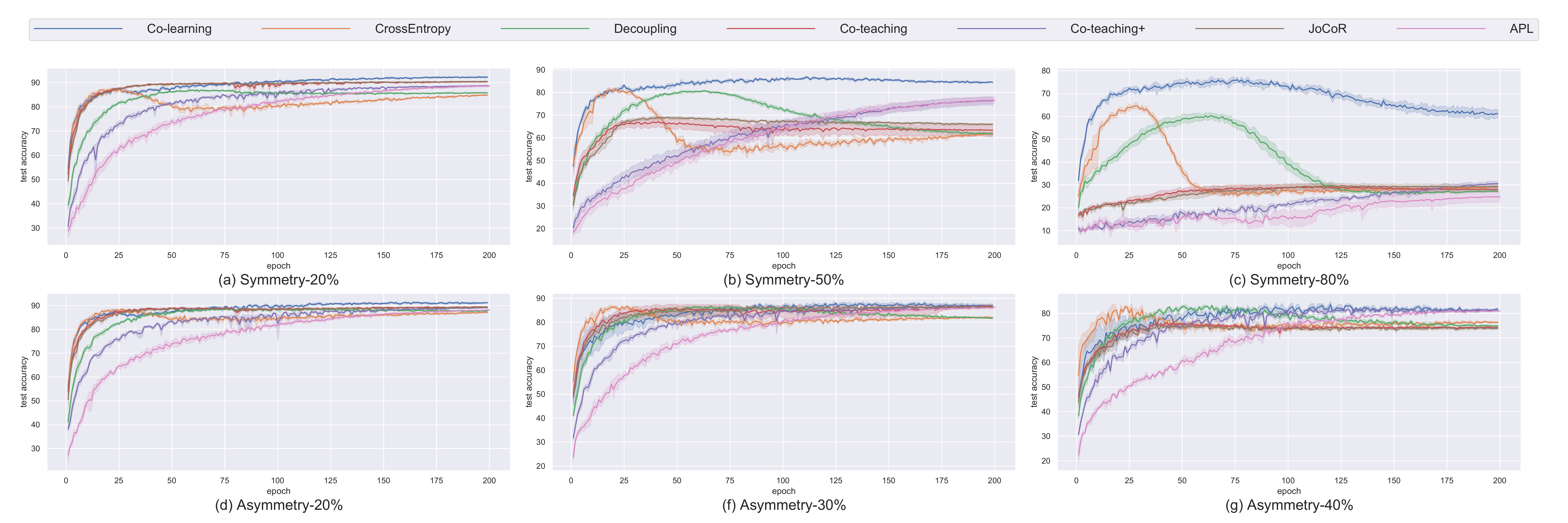} 
\caption{Results on CIFAR-10 dataset.}
\label{fig:cifar10} 
\end{figure*}

\begin{table*}[t]
\centering
\setlength{\tabcolsep}{3mm}{
\begin{tabular}{lrrrrrrr}
\toprule
Flipping-Rate    & Standard       & Decoupling     & Co-teaching    & Co-teaching+   & JoCoR          & APL & Co-learning \\
\midrule
Symmetric-20\%   & 84.81$\pm$0.24 & 85.75$\pm$0.31 & 90.29$\pm$0.19 & 88.63$\pm$0.32 & 90.43$\pm$0.25 & 88.54$\pm$0.45 & \textbf{92.21}$\pm$0.31\\
Symmetric-50\%   & 61.49$\pm$0.58 & 61.93$\pm$0.82 & 63.45$\pm$3.89 & 76.27$\pm$2.80 & 66.00$\pm$0.53 & 76.51$\pm$1.73 & \textbf{84.49}$\pm$0.34\\
Symmetric-80\%   & 28.98$\pm$0.26 & 27.23$\pm$0.84 & 28.03$\pm$1.67 & 30.37$\pm$1.69 & 29.19$\pm$1.64 & 24.75$\pm$2.87 & \textbf{61.20}$\pm$2.29      \\
Asymmetric-20\%  & 87.00$\pm$0.20 & 87.66$\pm$0.29 & 89.38$\pm$0.33 & 89.00$\pm$0.18 & 89.20$\pm$0.26 & 88.02$\pm$0.29 & \textbf{91.07}$\pm$0.32      \\
Asymmetric-30\%  & 81.99$\pm$0.31 & 81.83$\pm$0.26 & 86.58$\pm$1.32 & 86.22$\pm$0.26 & 86.41$\pm$0.45 & 86.03$\pm$0.21 & \textbf{86.89}$\pm$0.87      \\
Asymmetric-40\%  & 76.30$\pm$0.34 & 74.97$\pm$0.38 & 74.25$\pm$0.38 & 81.25$\pm$0.75 & 73.95$\pm$1.00 & 80.97$\pm$0.19 & \textbf{81.42}$\pm$0.52      \\
\bottomrule
\end{tabular}}
\caption{Average test accuracy (\%) on CIFAR-10 over the last 10 epochs.}
\label{tab:cifar10}
\end{table*}

\section{Experiments}
\subsection{Experimental settings}
\noindent{\bf Datasets.} Experiments are conducted on four benchmark datasets: CIFAR-10, CIFAR-100, Animal-10N \cite{song2019selfie}, and Food-101N \cite{lee2018cleannet} which are popularly used for evaluating models in learning from noisy labels~\cite{shu2019meta,arazo2019unsupervised}. 

\begin{figure}[htbp]
\centering
\includegraphics[width=0.45\textwidth]{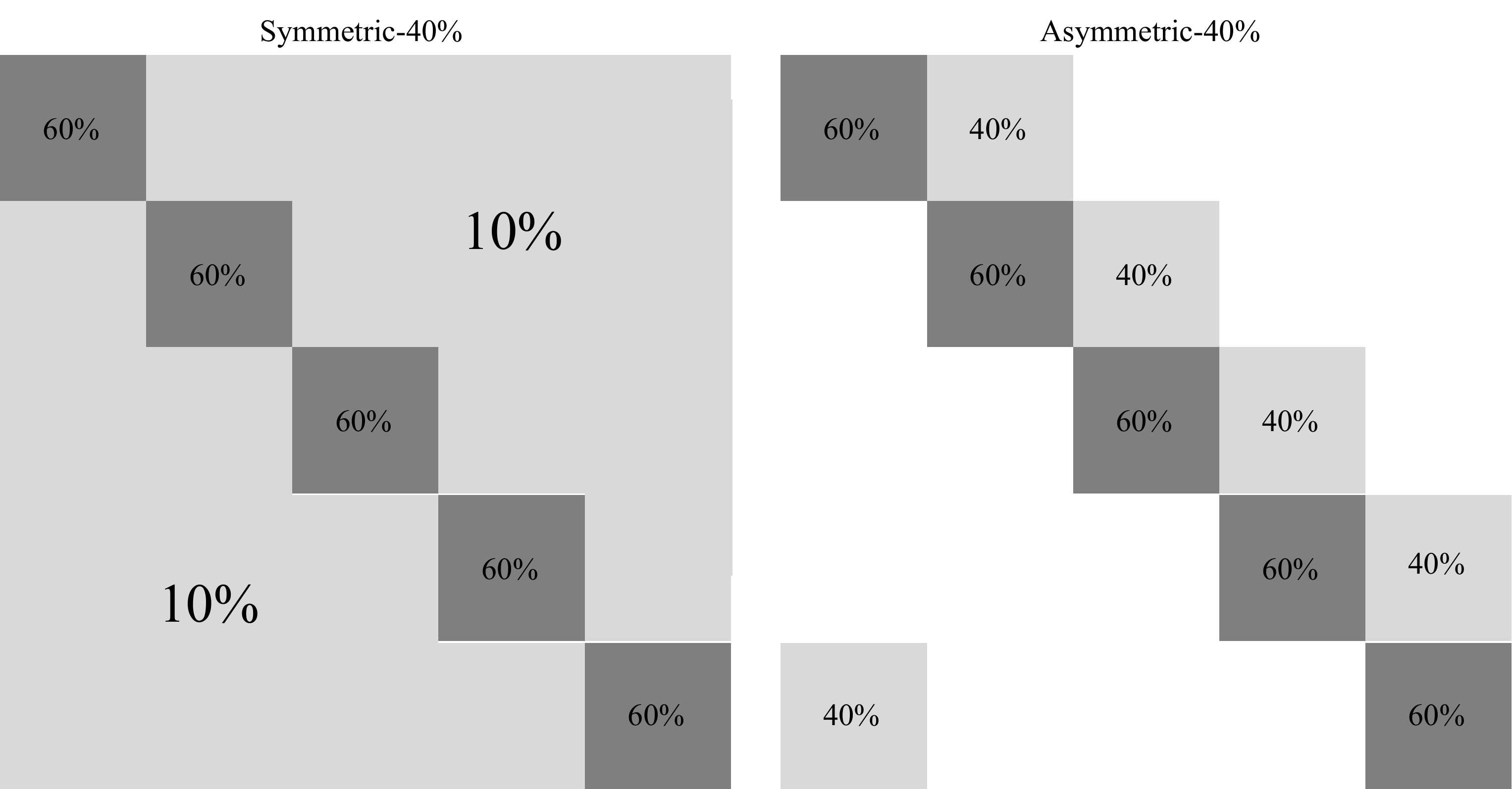} 
\caption{Example of noise transition matrices (taking 5 classes and noise ratio 0.4 as an example)}
\label{fig:noise_type} 
\end{figure}

Since CIFAR-10 and CIFAR-100 are clean datasets with true labels, we need to corrupt the original labels manually. Similar to previous works~\cite{reed2015training, patrini2017making, tanaka2018joint}, we consider two types of synthetic label noise: \emph{symmetric} noise and \emph{asymmetric} noise. Following \cite{zhang2018generalized,wang2019symmetric}, label noise is uniformly distributed among all categories with probability $\frac{p}{|C|}, p \in [0, 1]$ in the symmetric noise setup. In the asymmetric noise setup, labels are generated by flipping each class to a similar class or the next class circularly with probability $p$. We corrupt CIFAR-10 and CIFAR-100 datasets by the label transition matrix $Q$, where $Q_{ij} = \mathrm{Pr}[\hat{y} = j | y = i]$ given that noisy $\hat{y}$ is flipped from clean $y$. Figure \ref{fig:noise_type} shows an example of noise transition matrix on CIFAR-100. 

\begin{figure*}[ht]
\centering
\includegraphics[width=\textwidth]{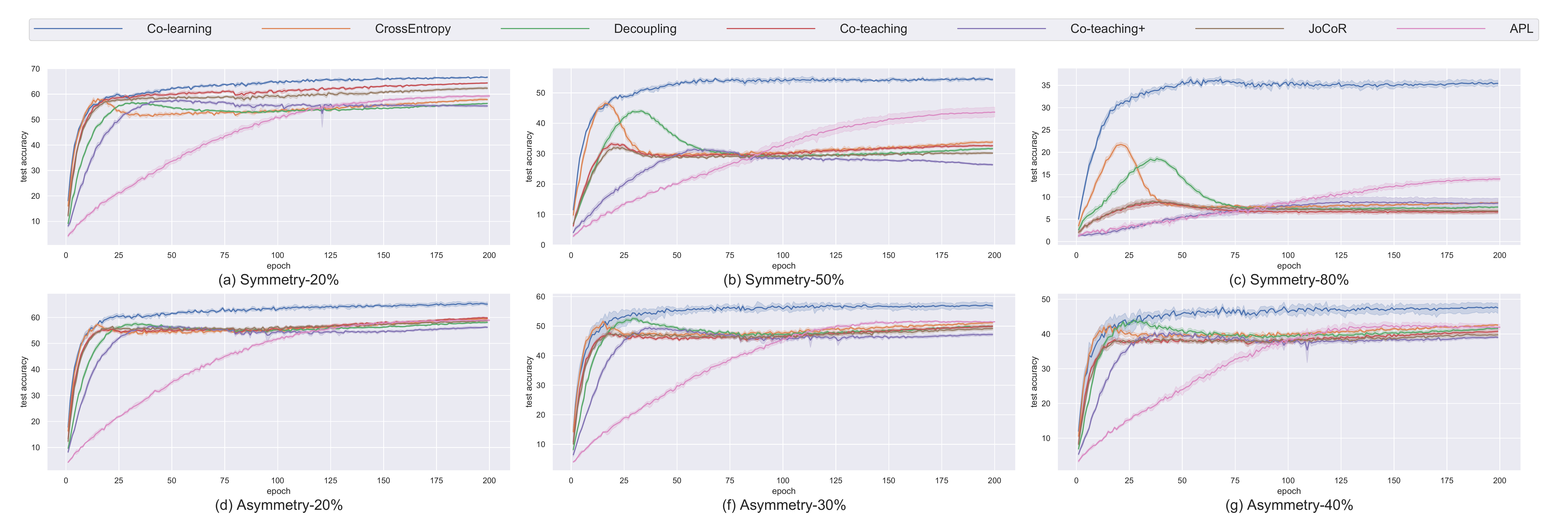} 
\caption{Results on CIFAR-100 dataset.}
\label{fig:cifar100} 
\end{figure*}

Following previous works \cite{wei2020combating,cheng2020learning}, we evaluate our method on various noise rates $\eta$, with $\eta \in \{20\%,50\%,80\%\}$ for symmetric label noise and $\eta \in \{20\%,30\%,40\%\}$ for asymmetric label noise. We use no more than 40\% because certain classes become theoretically indistinguishable for asymmetric noise larger than 50\%. As for the asymmetric noise of CIFAR-10, following \cite{patrini2017making} the noisy labels were generated by mapping "truck" $\rightarrow$ "automobile", "bird" $\rightarrow$ "airplane", "deer" $\rightarrow$ "horse" and "cat" $\rightarrow$ "dog" with probability $p$. But for the asymmetric noise of CIFAR-100, we generate labels by flipping each class to the next class with probability $p$.

Besides, Animal-10N \cite{song2019selfie} is a real-world noisy dataset of human-labeled online images for ten confusing animals, with 50,000 training and 5,000 testing images. As people may label images with confusing patterns, this dataset has natural noisy data. Its noise rate is estimated at 8\%. Food-101N is a webly noisy dataset that \cite{lee2018cleannet} collects 310k images from Google, Bing, Yelp, and TripAdvisor using the Food-101 \cite{bossard2014food} taxonomy. We train models on Food-101N and evaluate them on the test set of Food-101.

\noindent{\bf Baselines.} We compare our Co-learning with the following state-of-the-art methods, and we implement all methods with their default settings by PyTorch based on their open-source code. 

\setlength{\leftmargini}{6mm}
\setlength{\leftmarginii}{6mm}
\begin{enumerate}
\item Standard cross-entropy, which refers to train on noisy datasets directly with standard cross-entropy loss. This method is applied as a simple baseline.
\item Decoupling~\cite{malach2017decoupling}, which updates the parameters only using samples that have different predictions from two classiﬁers. 
\item Co-teaching~\cite{han2018co}, which trains two networks simultaneously and cross-updates parameters of their peer networks. 
\item Co-teaching+~\cite{yu2019does}, which trains two networks simultaneously and cross-updates their peer networks with the samples with different predictions from two classiﬁers. 
\item JoCoR~\cite{wei2020combating}, which trains two networks jointly and makes predictions from two classifiers similar by a Jensen-Shannon divergence.
\item APL~\cite{ma2020normalized}, which proposes a framework that combines two robust loss functions that mutually boost each other.
\end{enumerate}

\noindent{\bf Network Structure and Optimizer.} We utilize ResNet-18 for CIFAR-10 and CIFAR-100 datasets. For Animal-10N and Food-101N datasets, we utilize ResNet-34.

For experiments on CIFAR-10 and CIFAR-100, Adam optimizer is applied with an initial learning rate of 0.001 and a momentum of 0.9. The batch size is fixed as 128. For each trial, we run 200 epochs, and the learning rate is linearly decayed to zero from 80 to 200 epochs. All the experiments are repeated five times with different random seeds, and we report the average test accuracy of five trials for each experiment. For experiments on the real-world dataset Animal-10N and Food-101N, Adam optimizer is also applied with an initial learning rate of 0.001 and a momentum of 0.9. The batch size is set to 64. We run each trial for 100 epochs.

\noindent{\bf Measurement.} To evaluate the performance, the test accuracy is used, i.e., test accuracy = (\# of correct predictions) / (\# of the test). As the experiments of the synthetic noisy datasets are repeated five times, the MEAN and STD of the test accuracy are calculated. We have been highlighted the error bar for STD as a shade in each figure.

\subsection{Comparison with the State-of-the-Arts}

\paragraph{Results on CIFAR-10.} Figure \ref{fig:cifar10} shows the test accuracy on CIFAR-10. Co-learning performs best in all six cases. In these figures, test accuracy of the standard cross-entropy first reaches a very high level and then decreases gradually, which indicates the memorization effects of neural networks. The higher the noise rates, the more serious the problems become. A robust training method toward learning with noisy labels should avoid overfitting on noisy training data. 

The test accuracy of different algorithms is compared and reported in Table \ref{tab:cifar10}. In the easiest Symmetry-20\% case, all approaches work well except standard cross-entropy, demonstrating their robustness to low-level label noise. When it comes to higher noisy rates like Symmetry-50\% and Asymmetry-40\%, Decoupling starts to overfit to noisy data as standard cross-entropy does. In the hardest Symmetry-80\% case, Co-learning significantly outperforms other approaches and suffers less from noisy training data. Co-learning performs better than others in high noisy rates due to the feature-dependent information that comes from self-supervision.

\paragraph{Results on CIFAR-100.} Table \ref{tab:cifar100} shows results on CIFAR-100. As there are only 10 classes in CIFAR-10 but 100 classes in CIFAR-100, the overall accuracy is much lower than the previous. The memorization effects of neural networks are severe in standard cross-entropy as well as other approaches like Decoupling and Co-teaching+. As we can see, the observations are consistently the same as those for the CIFAR-10 dataset, which is, Co-learning still works much better than all other approaches.  In the easiest Symmetry-20\% case, only standard cross-entropy and Decoupling fail to combat noisy training data while others perform well. However, Co-learning works significantly better than other methods under high noisy rate cases such as Symmetry-50\% and Symmetry-80\%.

Figure \ref{fig:cifar100} shows the test accuracy during the training process with different noise types. Co-learning performs more stable learning curves with much less overfitting than other methods.

\begin{table*}[ht]
\centering
\setlength{\tabcolsep}{3mm}{
\begin{tabular}{lrrrrrrr}
\toprule
Flipping-Rate    & Standard       & Decoupling     & Co-teaching    & Co-teaching+   & JoCoR          & APL & Co-learning \\
\midrule
Symmetric-20\%   & 57.79$\pm$0.44 & 56.18$\pm$0.32 & 64.28$\pm$0.32 & 55.40$\pm$0.71 & 62.29$\pm$0.71 & 59.21$\pm$0.50 & \textbf{66.58}$\pm$0.15 \\
Symmetric-50\%   & 33.75$\pm$0.46 & 31.58$\pm$0.54 & 32.62$\pm$0.51 & 26.49$\pm$0.45 & 30.19$\pm$0.60 & 43.53$\pm$1.84 & \textbf{54.54}$\pm$0.43 \\
Symmetric-80\%   & 8.64$\pm$0.22 & 7.71$\pm$0.23 & 6.65$\pm$0.71 & 8.57$\pm$1.55 & 6.84$\pm$0.92 & 13.97$\pm$0.53 & \textbf{35.45}$\pm$0.79      \\
Asymmetric-20\%  & 59.36$\pm$0.36 & 57.97$\pm$0.24 & 59.76$\pm$0.53 & 56.11$\pm$0.60 & 58.58$\pm$0.51 & 58.89$\pm$0.40 & \textbf{65.26}$\pm$0.76 \\
Asymmetric-30\%  & 51.06$\pm$0.44 & 49.86$\pm$0.54 & 49.53$\pm$0.79 & 47.12$\pm$0.73 & 49.04$\pm$0.91 & 51.46$\pm$0.15 & \textbf{56.97}$\pm$1.22  \\
Asymmetric-40\%  & 42.49$\pm$0.23 & 41.51$\pm$0.67 & 40.62$\pm$0.79 & 38.98$\pm$0.54 & 39.72$\pm$0.76 & 41.96$\pm$0.92 & \textbf{47.62}$\pm$0.79  \\
\bottomrule
\end{tabular}}
\caption{Average test accuracy (\%) on CIFAR-100 over the last 10 epochs.}
\label{tab:cifar100}
\end{table*}

\begin{table}[ht]
\centering
\setlength{\tabcolsep}{8mm}{
\begin{tabular}{lrr}
\toprule
Methods      & best   & last   \\
\midrule
Standard CE  & 82.68  & 81.10 \\
Decoupling   & 79.22  & 78.24 \\
Co-teaching  & 82.43  & 81.52 \\
Co-teaching+ & 50.66  & 48.52 \\
JoCoR        & 82.82  & 81.06 \\
Co-learning  & \textbf{82.95} & \textbf{82.18} \\
\bottomrule
\end{tabular}
}
\caption{Classification accuracy (\%) on Animal-10N dataset.}
\label{tab:animal}
\end{table}

\paragraph{Results on Animal-10N} Table \ref{tab:animal} summarizes the test results on the real-world noisy labels using the Animal-10N dataset, where \textit{best} denotes the optimal validation accuracy, and \textit{last} denotes the final test accuracy. As Co-teaching, Co-teaching+, and JoCoR require a noisy rate estimation to assist their training, we roughly set it as 10\%. Despite training with less prior knowledge, Co-learning consistently outperforms other methods.

\paragraph{Results on Food-101N} Table \ref{tab:food} illustrates the test results on Food-101N dataset, where \textit{best} denotes the optimal validation accuracy, and \textit{last} denotes the final test accuracy. Co-learning still outperforms other methods on the webly real-world noisy dataset, which demonstrates its robustness.

\begin{table}[ht]
\centering
\setlength{\tabcolsep}{8mm}{
\begin{tabular}{lrr}
\toprule
Methods      & best   & last   \\
\midrule
Standard CE  & 84.50  & 83.86 \\
Decoupling   & 85.53  & 85.28 \\
Co-teaching  & 61.91  & 61.86 \\
Co-teaching+ & 81.61  & 81.24 \\
JoCoR        & 77.94  & 77.86 \\
Co-learning  & \textbf{87.57}  & \textbf{86.56} \\
\bottomrule
\end{tabular}
}
\caption{Classification accuracy (\%) on Food-101N dataset.}
\label{tab:food}
\end{table}

\subsection{Ablation study}

\begin{figure}[htbp]
\centering
\includegraphics[width=0.40\textwidth]{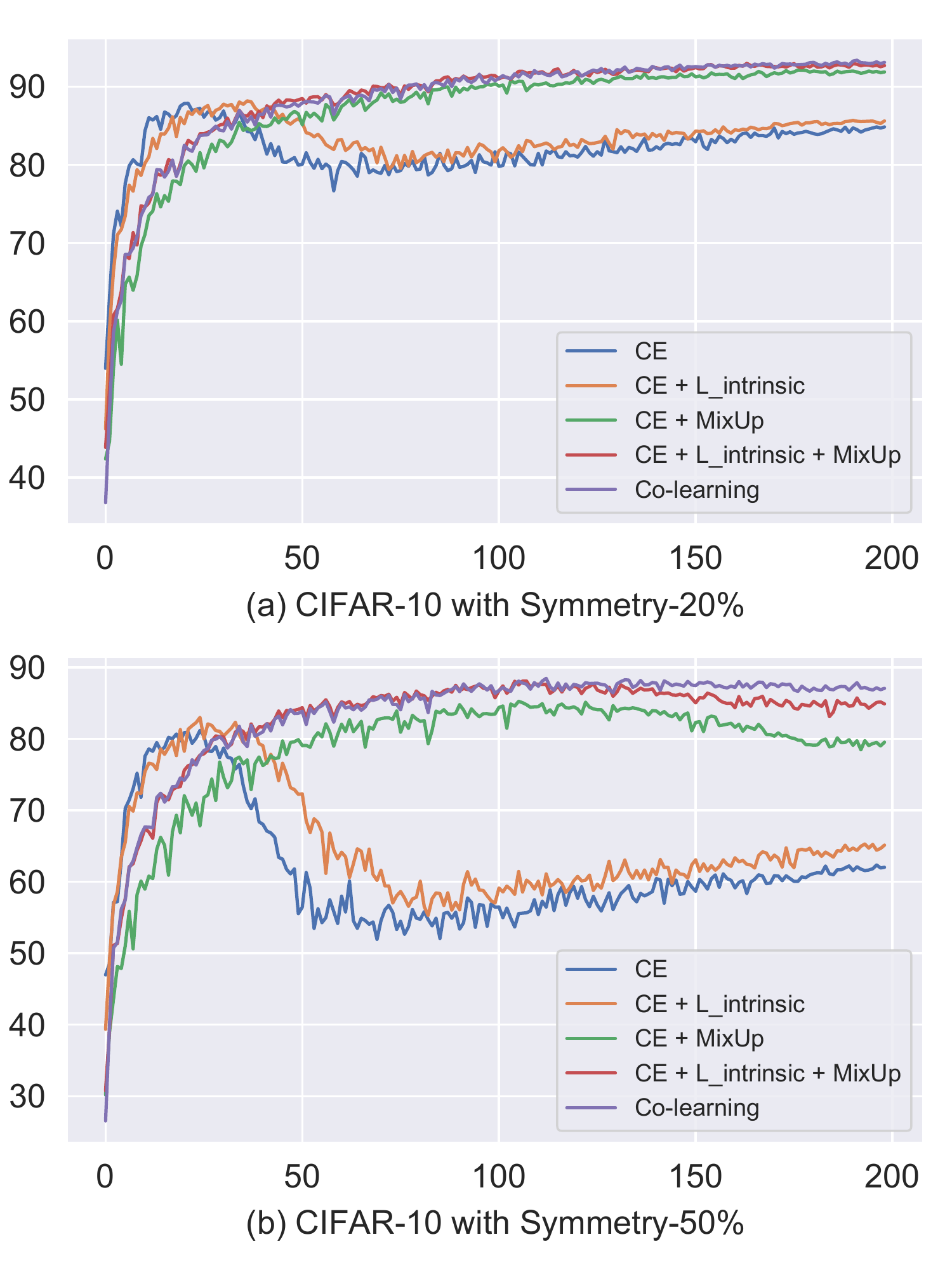} 
\caption{Results of ablation study on CIFAR-10 with Symmetry-20\% and Symmetry-50\% noise.}
\label{fig:ablation_study} 
\end{figure}

To study the effect of our proposed components, the ablation study is conducted on CIFAR-10 with Symmetry-20\% and Symmetry-50\% noise, as shown in Figure \ref{fig:ablation_study}. 

Standard cross-entropy suffers from overfitting on noisy training data. Simply adding a self-supervision signal helps to a small extent but is not good enough. The main issue is the slow convergence of self-supervised learning, making the feature encoder $f$ quickly overfits label-dependent information but ignores feature-dependent information. 

In order to slow down the inaccurate convergence of supervised learning, a data augmentation method, MixUp, is utilized to avoid early overfitting. As we can see, a balanced training that prompts both label-dependent information and feature-dependent information play crucial roles during the training procedure. Though Co-learning without structural similarity loss can obtain a good performance under the low noisy rate Symmetry-20\%, it still overfits noisy training data under the high noisy rate Symmetry-50\%. Co-learning utilizes the structural similarity between different pretext tasks to regularize supervised learning and keeps a robust noise-tolerant training procedure. These results show that self-supervised learning can help combat noisy labels, and a balanced training manner further regularizes the model from overfitting matters. Our proposed Co-learning is a more efficient paradigm to take advantage of feature-dependent information.

\subsection{Discussion}

Though these co-training-based methods for combating noisy labels are based on the assumption that different views of data should provide extra valuable information, networks with the same architecture provide limited variety caused by random initialization. In general, Decoupling firstly leverages the co-training method with a cross-update technique based on disagreements from two different views. Inspired by the small-loss trick from MentorNet, Co-teaching cross-updates samples selected by the small-loss. Co-teaching+ combines disagreement strategy and the small-loss trick, which means add a disagreement selection step in Co-teaching. JoCoR simplifies the above ideas and jointly updates two networks still based on the small-loss trick and an extra Jensen-Shannon divergence between predictions from two peer networks.

Co-learning further predigests these co-training methods through a shared feature encoder with two exclusive heads, which provides different views of data. Besides, Co-learning explores a reasonable scheme that leverages unsupervised learning to assist robust deep learning.

\section{Conclusion}

In this paper, we propose Co-learning for combating noisy labels by utilizing both label-dependent information from supervised learning and feature-dependent information from self-supervised learning. Instead of training two classifiers, Co-learning trains two heads with a shared feature encoder and regularizes the model with both the intrinsic similarity and the structural similarity. Co-learning achieves robust training without trivial tricks, while other co-training-based methods need the small-loss trick and noisy rate estimation. Through extensive experiments on synthetic and real-world noisy datasets, we demonstrate that Co-learning outperforms other co-training-based methods and provides a new perspective to leverage self-supervised learning to assist robust deep learning. In the future, we will further explore the theoretical foundation and generalization analysis of Co-learning.

\balance

\bibliographystyle{ACM-Reference-Format}
\bibliography{ref}

\appendix
\newpage

\section{Implementation details}

\subsection{Dataset}

The detailed information about datasets used in this paper is shown in Table \ref{tab:datasets}.

\begin{table}[h]
\centering
\setlength{\tabcolsep}{2mm}{
\begin{tabular}{lrrrr}
\toprule
Datasets      & \# of training   & \# of test & \# of class & size   \\
\midrule
CIFAR-10   & 50,000   & 10,000 & 10  & 32 $\times$ 32 \\
CIFAR-100  & 50,000   & 10,000 & 100 & 32 $\times$ 32 \\
Animal-10N & 50, 000  & 5,000  & 10  & 64 $\times$ 64   \\
Food-101N & 310,000   & 5,000  & 101 & 128 $\times$ 128 \\
\bottomrule
\end{tabular}
}
\caption{Summary of datasets used in the this paper.}
\label{tab:datasets}
\end{table}

\subsection{Transformation Details}

The detailed strong transformation $\mathcal{T}$ and weak transformation $\mathcal{T}'$ are illustrated in PyTorch-style notations below.

For $\mathcal{T}$, geometric transformations are \texttt{RandomResizedCrop} with scale in [0.08, 1.0], and \texttt{RandomHorizontalFlip} with probability of 0.5. Color transformations are \texttt{ColorJitter} with {brightness, contrast, saturation, hue} strength of \{0.4, 0.4, 0.4, 0.1\} with probability of 0.8, and \texttt{RandomGrayscale} with probability of 0.2.

% \paragraph{Weak transformation}

While the strong transformation $\mathcal{T}$ includes geometric transformations and color transformations, the weak transformation $\mathcal{T}'$ only consists of geometric transformations, i.e. \texttt{RandomResizedCrop} with scale in [0.08, 1.0], and \texttt{RandomHorizontalFlip} with probability of 0.5.

\subsection{Training details}

We implement co-training-based noisy label learning methods into a unified framework for fair comparisons. ResNet-18 is utilized for CIFAR-10 and CIFAR-100 datasets. Adam optimizer (momentum=0.9) is utilized with an initial learning rate of 0.001, and the batch size is set to 128. We run 200 epochs and linearly decay the learning rate to zero from 80 to 200 epochs.  The hyperparameter settings on CIFAR-10 and CIFAR-100 datasets are the same as their original papers. Specially, for Co-teaching and Co-teaching+, we set $num\_gradual = 1$; for JoCoR, we set $num\_gradual = 1, co\_lambda = 0.1$. 

\section{Additional Experiments}

\subsection{Comparison with loss-design methods}

\begin{figure}[htb]
\centering
\includegraphics[width=0.46\textwidth]{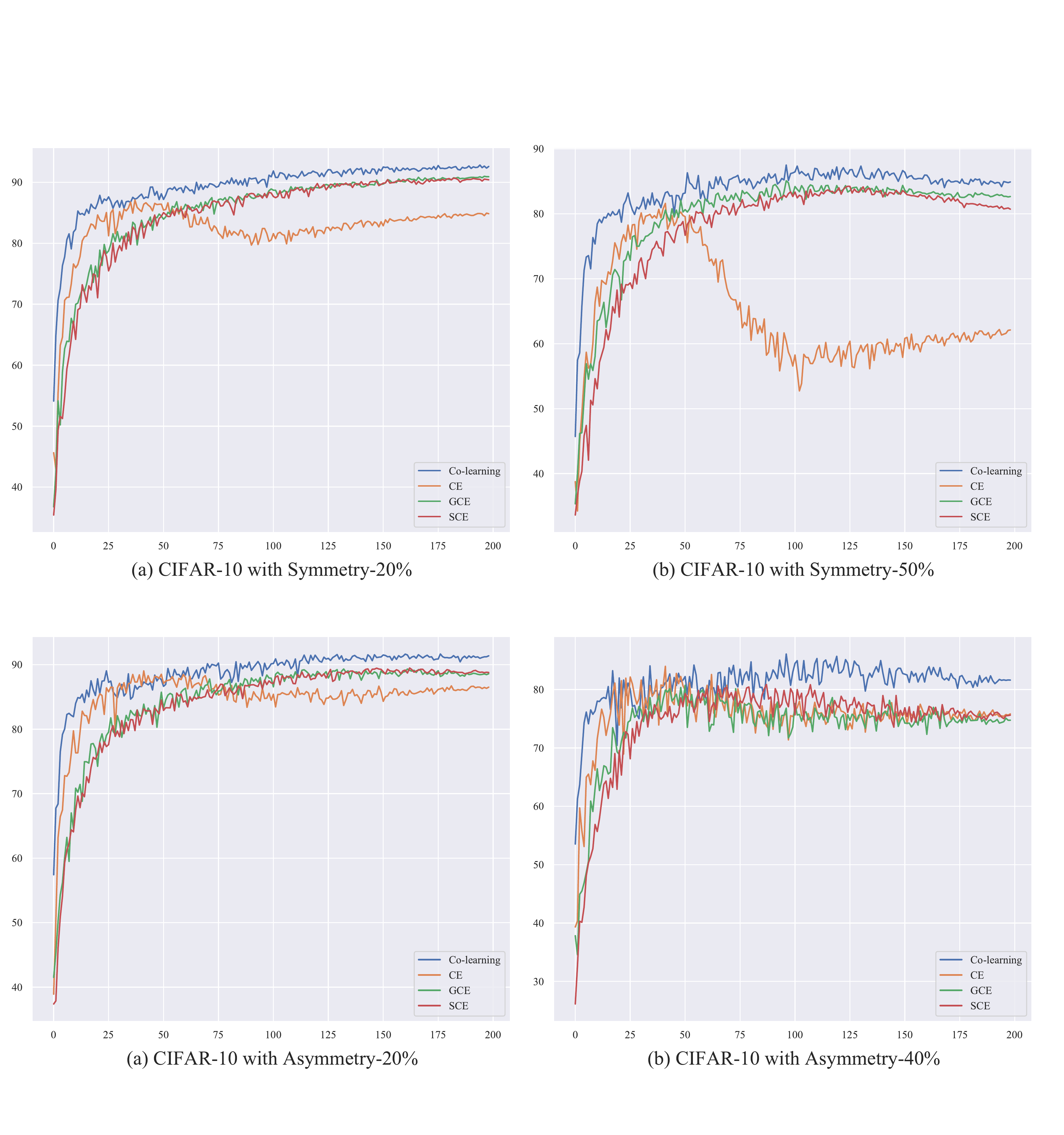} 
\caption{Comparison between Co-learning and other loss-design methods on CIFAR-10.}
\label{fig:additional_loss_design} 
\end{figure}

Figure~\ref{fig:additional_loss_design} shows the comparison between Co-learning and noisy learning methods based on loss-design, known as GCE and SCE. Co-learning converges faster than other methods under all these four conditions with the highest accuracy, though GCE and SCE have smoother convergence curves.

\subsection{MixUp v.s. Weighted supervised loss}

\begin{figure}[htbp]
\centering
\includegraphics[width=0.42\textwidth]{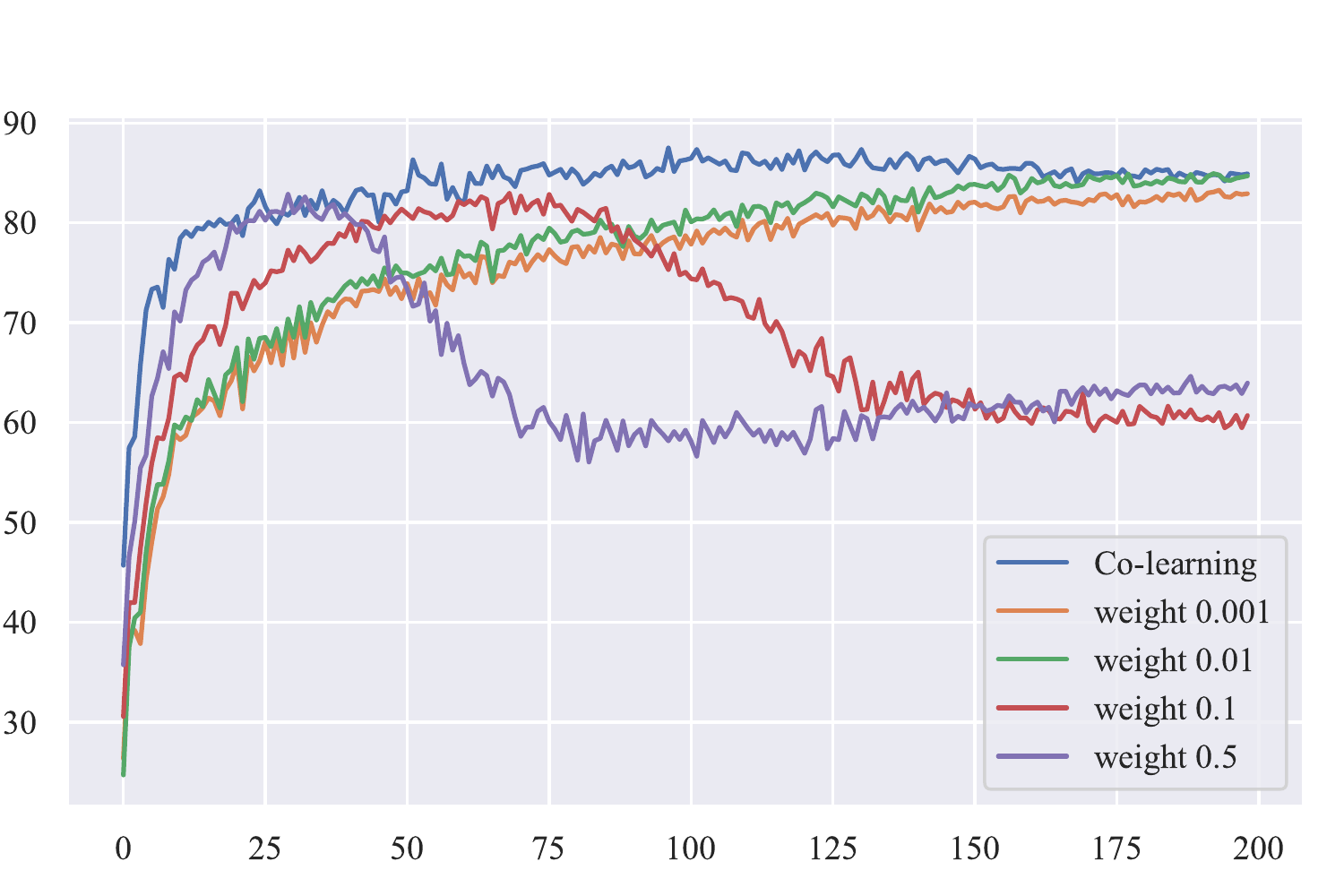} 
\caption{Co-learning with weighted supervised loss.}
\label{fig:additional_weight} 
\end{figure}

Co-learing adopts MixUp during the training to slow down the overfitting on noisy labels. However, figure~\ref{fig:additional_weight} shows that replacing MixUp by multiplying small weights on supervised loss can achieve similar results. Supervised loss with the weight of 0.001 and weight 0.01 have smooth convergence curves, while larger weights fail. We prefer the MixUp version as it is more insensitive to hyperparameters.

\end{document}